\documentclass[11pt]{article}

\usepackage[final]{acl}

\usepackage{times}
\usepackage{latexsym}

\usepackage[T1]{fontenc}
\usepackage[utf8]{inputenc}

\usepackage{microtype}

\usepackage{inconsolata}
\usepackage{CJKutf8}
\usepackage{graphicx}
\usepackage{amsmath, bm, amssymb}
\usepackage{multirow}
\usepackage{booktabs}
\usepackage{array}
\usepackage{amssymb}
\title{Dynamic Expert Specialization: Towards Catastrophic Forgetting-Free Multi-Domain MoE Adaptation}

\author{
Junzhuo Li\textsuperscript{\dag}\textsuperscript{\ddag}, 
Bo Wang\textsuperscript{\dag}, 
Xiuze Zhou\textsuperscript{\dag}\textsuperscript{\ddag}, 
\and 
Xuming Hu\textsuperscript{\dag}\textsuperscript{\ddag}\thanks{Corresponding author} \\
\textsuperscript{\dag}The Hong Kong University of Science and Technology (Guangzhou) \\
\textsuperscript{\ddag}The Hong Kong University of Science and Technology \\
\texttt{\{jz.li, bo.wang, xz.zhou\}@connect.hkust-gz.edu.cn} \\
\texttt{xuminghu@hkust-gz.edu.cn}
}
\begin{document}
\maketitle
\begin{abstract}
Mixture-of-Experts (MoE) models offer immense capacity via sparsely gated expert subnetworks, yet adapting them to multiple domains without catastrophic forgetting remains an open challenge. Existing approaches either incur prohibitive computation, suffer cross-domain interference, or require separate runs per domain. We propose DES-MoE, a dynamic expert specialization framework for multi-domain adaptation of Mixture-of-Experts models. DES-MoE addresses catastrophic forgetting through three innovations: (1) an adaptive router balancing pre-trained knowledge retention and task-specific updates via distillation, (2) real-time expert-domain correlation mapping to isolate domain-specific gradients, and (3) a three-phase adaptive fine-tuning schedule that progressively freezes non-specialized parameters. Evaluated on six domains (math, code, law, etc.), DES-MoE matches single-domain ESFT performance while training one unified model, reduces forgetting by 89\% compared to full fine-tuning as domains scale from 2 to 6, and achieves 68\% faster convergence than conventional methods. Our work establishes dynamic expert isolation as a scalable paradigm for multi-task MoE adaptation.
\end{abstract}

\section{Introduction}

Mixture-of-Experts (MoE) models have emerged as a promising architecture for scaling up deep learning, especially for large language models \citep{albert-2024-mixtral, dai-etal-2024-deepseekmoe, xue2024openmoe, qwen_moe, sun2024hunyuanlargeopensourcemoemodel}. By using a sparsely-gated routing mechanism, MoEs activate only a small subset of “expert” subnetworks for each input, dramatically increasing model capacity without proportional increases in computation. This approach has enabled models with hundreds of billions to trillions of parameters. 

However, adapting MoE models to new domains or tasks remains challenging. Sparse MoE architectures can suffer degraded performance under distribution shifts. A given domain might activate a particular subset of experts heavily, and different domains tend to rely on different experts \citep{li-etal-2025-decoding}. Naïvely fine-tuning a MoE on multi-domain data can therefore lead to inefficiencies: some experts may be over-specialized to certain domains while others are under-utilized, causing unstable training and suboptimal generalization. The hard routing decisions in MoEs, while efficient, also mean that mistakes in the routing (or changes in domain characteristics) can significantly impact performance on a new domain if not properly addressed.

Recent research has begun exploring methods to efficiently fine-tune MoEs for downstream tasks. One notable approach is Expert-Specialized Fine-Tuning (ESFT) \citep{wang2024let}, which adapts an MoE by updating only the experts most relevant to a target task or domain while freezing the rest show that by tuning a small subset of experts selected for a specific task, one can match or even exceed the performance of full model fine-tuning, with substantially improved efficiency. This static, task-specific strategy validates the intuition that different experts encode different knowledge and that focusing on the most pertinent experts can yield efficient adaptation. Yet, it also highlights a fundamental limitation: the approach is inherently tied to a single domain or task at a time. For each new domain, one must determine a new expert subset and fine-tune again from scratch, which is inefficient and poorly scalable as the number of domains grows. Moreover, static expert allocation fails to exploit commonalities between domains; experts tuned for one domain are not easily reused for another, even if some knowledge could be shared.

In this paper, we present \textbf{D}ynamic \textbf{E}xpert \textbf{S}pecialization (DES-MoE), a lightweight multi-domain fine-tuning framework for Mixture-of-Experts models. DES-MoE combines a learnable adaptive router with online expert–domain correlation tracking to dynamically select and sparsely update only the most relevant experts on a per-batch basis. Fine-tuning proceeds through three progressive stages—Warm-Up, Stabilization, and Consolidation—each imposing stricter masks on router, backbone, and expert parameters to (1) rapidly discover domain signals, (2) refine gating and minimize cross-domain interference, and (3) lock in specialized adaptations with parameter-efficient updates.

\paragraph{Contributions} Our main contributions are as follows:
\begin{itemize}
\item \textbf{Dynamic multi-domain routing}
A unified fine-tuning framework with a learnable, per-input router that replaces static expert assignments and adapts to heterogeneous domains in a single MoE model.
\item \textbf{Expert–domain correlation and sparse updates}
An online tracking mechanism that identifies the most relevant experts for each domain batch and restricts fine-tuning to that subset, cutting compute and preventing negative transfer.
\item \textbf{Progressive three-phase specialization schedule}
A parameter-masking regimen (Warm-Up, Stabilization, Consolidation) that gradually freezes router, backbone, and non-selected experts to stabilize training and concentrate final updates on domain-specific experts.
\end{itemize}

% We have shown that DES-MoE isolates domain-specific parameters, prevents gradient interference, and dramatically reduces compute. Empirically on both specialized tasks (math, code, NLU, summarization, legal reasoning, translation) and broad benchmarks, a single DES-MoE model matches or outperforms separate single-domain fine-tuning while preserving general capabilities as domains scale. Moreover, DES-MoE cuts fine-tuning time by more than two-thirds compared to full-parameter updates.

Empirically, DES-MoE matches or exceeds separate single-domain fine-tuning on six specialized tasks and preserves general benchmarks as the number of domains grows, all while reducing total fine-tuning time by over two-thirds compared to full-parameter updates.

\section{Related Work}

\paragraph{Parameter-Efficient Fine-Tuning}
Parameter‐efficient fine‐tuning (PEFT) has become a popular means to adapt large language models to downstream tasks with minimal extra training cost, and existing methods for dense architectures fall into three camps: augment‐and‐freeze approaches that insert and train only a small set of new parameters—such as prompt tuning \citep{lester2021power}, prefix tuning \citep{li2021prefix, liu2021p} and adapters \citep{houlsby2019parameter, pfeiffer2020adapterfusion, he2021towards,wang2022adamix}
—selective fine‐tuning that updates only a subset of existing weights \citep{guo2020parameter, gheini2021cross, he2023sensitivity, vucetic2022efficient, liao2023parameter, ansell-etal-2022-composable, sung2021training, xu2021raise}
, and low‐rank adaptation techniques like LoRA \cite{hu2022lora} and its many refinements \citep{zhang2023adalora, ding2023sparse, lin2024lora, liu2023moelora, dou-etal-2024-loramoe, pan2024lisa}%(Zhang et al., 2023a; Ding et al., 2023; Lin et al., 2024; Liu et al., 2023; Dou et al., 2024). 

\paragraph{Mixture of Experts}

Mixture-of-Experts (MoE) architectures \citep{fedus-2021-switch, lepikhin2021gshard, zoph2022st, dai-etal-2024-deepseekmoe} have demonstrated that one can decouple computational cost from parameter count by selectively activating only a subset of ``experts'' for each input \citep{fedus-2021-switch, lepikhin2021gshard, roller2021hash, dai-etal-2022-stablemoe,xue2024openmoe, li-etal-2025-decoding}. 
More recently, research has shifted from coarse-grained MoE \citep{albert-2024-mixtral} designs—characterized by a small number of high-dimensional experts—to fine-grained \citep{ludziejewski2024scaling, dai-etal-2024-deepseekmoe, liu2024deepseek, deepseekai2024deepseekv3technicalreport} configurations with many low-dimensional experts, allowing for more precise expert selection and highly efficient task-specific tuning.

% However, extending PEFT to sparse Mixture‐of‐Experts models remains underexplored—\citet{wang2024let} introduce ESFT to backpropagate only through experts most relevant to a single downstream task, yet it cannot handle mixed‐domain fine‐tuning—and in this work we propose DES-MoE that selects and updates experts by downstream‐task affinity to enable efficient, multi‐domain adaptation in sparse MoE architectures.
However, extending PEFT to sparse Mixture‐of‐Experts models remains underexplored. For instance, \citet{wang2024let} introduce ESFT, which fine-tunes only the experts most relevant to a given downstream task based on expert–domain affinity. This design improves task performance while mitigating catastrophic forgetting, but its major limitation lies in the requirement to train a separate model for each domain, leading to prohibitive computational and storage costs. In contrast, our work proposes \textbf{DES-MoE}, a \textit{dynamic} expert specialization fine-tuning framework that adaptively selects and updates experts according to downstream task affinity. This enables efficient \emph{multi-domain} adaptation in sparse MoE architectures without the need for task-specific model duplication.

\begin{figure*}[t]
  \includegraphics[width=\textwidth]{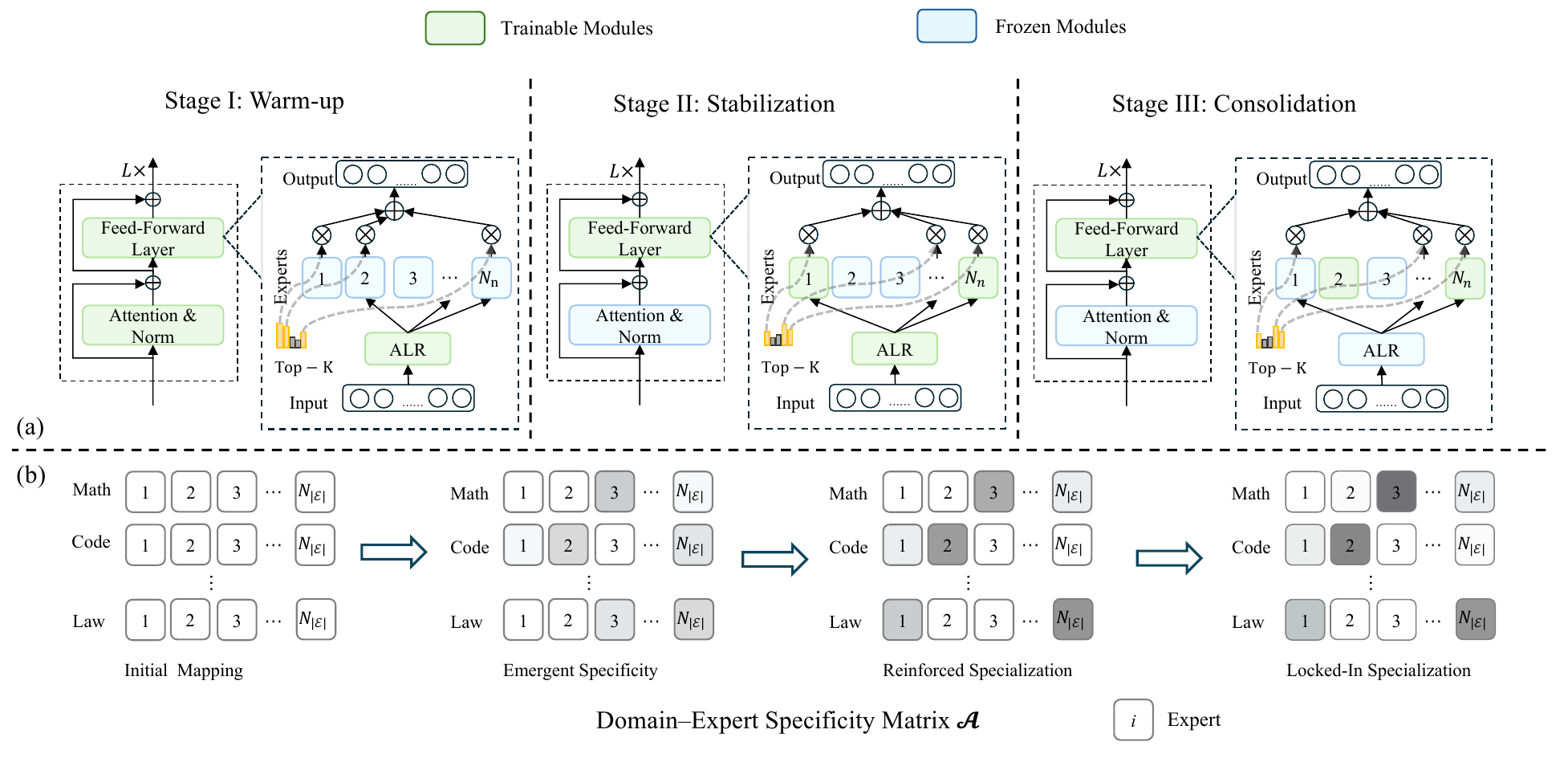}
  \caption{The DES-MoE framework. (a) Progressive Parameter Specialization Schedule: in \textbf{Stage I (Warm-up)} all router and expert parameters are trainable; in \textbf{Stage II (Stabilization)} only the adaptive router and domain-relevant experts are updated; in \textbf{Stage III (Consolidation)} the router and unrelated experts are frozen, and only the final domain-specific experts remain trainable. (b) Evolution of the Domain–Expert Specificity Matrix $\mathcal{A}$: each cell’s shading indicates how strongly an expert is associated with a given domain, progressing from a uniform (blank) mapping, through emergent and reinforced specialization, to a locked-in final mapping. }%Full fine‐tuning (FFT) suffers an accelerating drop from 61.7 to 55.5, LoRA and static ESFT decline more moderately (ending at 56.7 and 58.7, respectively), while DES‐MoE remains effectively flat (62.5→62.4), demonstrating its ability to preserve general knowledge under multi‐domain adaptation.}
  \label{fig:framework}
\end{figure*}

\section{DES-MoE}
Figure~\ref{fig:framework} provides an overview of the DES-MoE framework. In this section, we detail its three core components. First, we describe the \textbf{Adaptive Lightweight Router} (ALR) (\S~\ref{subsec:adaptive_lightweight}), which replaces the frozen pre-trained gating layer with a small MLP trained via a combined task and distillation objective. Next, we introduce the \textbf{Domain-Guided Expert Specialization} (DGES) (\S~\ref{subsec:expert_specialization}), a mechanism for dynamically identifying which experts are most relevant to each domain and applying selective gradient masking. Finally, we present the \textbf{Progressive Parameter Specialization Schedule} (\S~\ref{subsec:progressive_parameter}), a three-phase training regimen that gradually freezes unrelated parameters to consolidate domain-specific expertise without disrupting the model’s general capabilities.

\subsection{Adaptive Lightweight Router}
\label{subsec:adaptive_lightweight}
Static routing mechanisms learned during pretraining often struggle to accommodate domain shifts encountered in multi-task fine-tuning. On one hand, updating the original router in full can overwrite valuable pretrained knowledge; on the other, freezing it entirely prevents the model from adapting to new domains. To strike a balance between knowledge preservation and domain adaptation, we introduce an \textbf{A}daptive \textbf{L}ightweight \textbf{R}outer (ALR) that augments the pretrained linear router with a shallow, trainable MLP and is trained under a dual-signal paradigm.

Concretely, given a token representation $\bm{h}_t \in \mathbb{R}^d$, we define the adaptive router
\begin{equation}
    R_{\text{adapt}}(\bm{h}_t) = \bm{W}_2\,\mathrm{GELU}\bigl(\bm{W}_1 \bm{h}_t + \bm{b}_1\bigr) + \bm{b}_2,
\end{equation}
where $\bm{W}_1\in\mathbb{R}^{d\times 4d}$, $\bm{b}_1\in\mathbb{R}^{4d}$ constitute the hidden layer, and $\bm{W}_2\in\mathbb{R}^{4d\times |\mathcal{E}|}$, $\bm{b}_2\in\mathbb{R}^{|\mathcal{E}|}$ project to the expert logits (with $|\mathcal{E}|$ being the total number of experts). We initialize $\bm{W}_2$ by copying the pretrained router’s weights and apply Kaiming initialization \citep{he2015devling} to $\bm{W}_1$, thus preserving the router’s original behavior at the start of fine-tuning. Let \(\bm{\theta}_{\text{router}} = \{\bm{W}_1, \bm{b}_1, \bm{W}_2, \bm{b}_2\}\) denote the set of trainable parameters of this adaptive router.

Training proceeds with two complementary loss components. First, a knowledge distillation loss:
% \begin{equation}
%     \mathcal{L}_{\mathrm{KD}} \;=\; \frac{1}{T}\sum_{t=1}^T \mathrm{KL}\Bigl(\sigma\bigl(R_{\text{orig}}(h_t)/\tau\bigr)\,\Vert\,\sigma\bigl(R_{\text{adapt}}(h_t)/\tau\bigr)\Bigr),
% \end{equation}
\begin{equation}
\begin{aligned}
\mathcal{L}_{\mathrm{KD}} = & \frac{1}{T}\sum_{t=1}^T \mathrm{KL}\Bigl( \\
& \sigma\bigl(R_{\text{orig}}(\bm{h}_t)/\tau\bigr)\Vert\sigma\bigl(R_{\text{adapt}}(\bm{h}_t)/\tau\bigr)\Bigr),
\end{aligned}
\end{equation}
encourages the adaptive router to mimic the pretrained routing patterns (with temperature $\tau=0.7$). Second, a task adaptation loss
\begin{equation}
    \mathcal{L}_{\mathrm{task}} =-\sum_{t=1}^T \log P\bigl(\bm{y}_t \mid \bm{h}_t, \bm{\theta}_{\text{router}}\bigr)
\end{equation}
drives specialization toward downstream objectives. We combine these via a time-dependent weighting,
\begin{equation}
    \mathcal{L}_{\text{router}} \;=\; \lambda(\alpha)\,\mathcal{L}_{\mathrm{KD}} \;+\;\bigl(1-\lambda(\alpha)\bigr)\,\mathcal{L}_{\mathrm{task}},
\end{equation}
where $\lambda(\alpha)=\max\bigl(0,\;1-\alpha\bigr)$, $\alpha\in[0,1]$ denotes the fraction of fine-tuning completed. Early in training ($\lambda\approx1$), the router remains close to its pretrained state; during the middle phase ($0.2<\lambda<0.7$), it gradually learns domain-specific routing preferences; and by the end ($\lambda=0$), it fully optimizes for task performance. This phased adaptation ensures a smooth and controlled transition from general pretrained knowledge to specialized behavior.

\subsection{Domain-Guided Expert Specialization}
\label{subsec:expert_specialization}
Experts trained jointly on multiple domains often suffer from destructive interference, as gradient updates from one domain can overwrite useful knowledge learned for another. To mitigate this, we introduce a \textbf{D}omain-\textbf{G}uided \textbf{E}xpert \textbf{S}pecialization (DGES) scheme that (i) uncovers each domain’s preferred experts, (ii) restricts updates to those experts, and (iii) preserves a pool of universally shared experts for cross-domain transfer.

Let the training data be partitioned by domain $\mathcal{D} = \bigcup_{d=1}^D \mathcal{D}_d$, with $N_d=|\mathcal{D}_d|$.  During an initial warmup phase, we record how often each expert $e$ is selected for inputs from domain $d$:
\begin{equation}
    A_d^{(e)} \;=\; \frac{1}{N_d} \sum_{(\bm{h}_t,\bm{y}_t)\in\mathcal{D}_d}
\mathbb{I}\bigl(e \in \mathrm{TopK}\!\bigl(R(\bm{h}_t)\bigr),
\end{equation}
where $(h_t, y_t)\in\mathcal{D}_d$ is a training pair from domain $d$, %where $\bm{h}_t$ is the model’s hidden representation at time step $t$ and $\bm{y}_t$ is the target, 
and $\mathbb{I}$ is the indicator function and $K$ is the number of experts routed per token.  Intuitively, $A_d^{(e)}$ captures the affinity between domain $d$ and expert $e$.

We then define a binary specialization matrix $\mathcal{M}\in\{0,1\}^{|\mathcal{D}|\times |\mathcal{E}|}$ by thresholding each row of $\bm{A}$:
\begin{equation}
    \mathcal{M}_{d,e} = 
\begin{cases}
1, & A_d^{(e)} \;\ge\;\phi\cdot\max_{e'}(A_d^{(e')}),\\
0, & \text{otherwise},
\end{cases}
\end{equation}
with relative threshold $\phi=0.6$.  During subsequent fine-tuning phases, when processing a batch drawn from domain $d$, we mask expert parameters so that only $\{\theta_e\!\mid\,M_{d,e}=1\}$ receive nonzero gradients:
\begin{equation}
    \frac{\partial\mathcal{L}}{\partial\theta_e}
=
\begin{cases}
\frac{\partial\mathcal{L}_{\mathrm{task}}}{\partial\theta_e}, & M_{d,e}=1,\\
0, & \text{otherwise}.
\end{cases}
\end{equation}

To avoid cross-domain conflict in mixed-domain batches, we either group examples by domain or, if necessary, apply this masking at the token level:
\begin{equation}
    g_e^{(t)} \;=\;
\sum_{i=1}^B \mathbb{I}\bigl(e\in \mathcal{E}_i\bigr)\,\frac{\partial \mathcal{L}_i}{\partial \theta_e},
\end{equation}
where $\mathcal{E}_i$ is the set of experts selected for token $i$ with domain label $d_i$. In other words, $g_e^{(t)}$ sums up only those per-token gradients for which expert $e$ actually participated, ensuring that experts masked out for a given token (because $M_{d_i,e}=0$) contribute nothing to its update.

% Finally, we periodically update $A_d^{(e)}$ with momentum,
The specialization matrix $M$ is periodically updated every $T_\text{update}$ steps:
\begin{equation}
    \hat{A}_d^{(e)} \;\leftarrow\; \alpha\,A_d^{(e)} + (1-\alpha)\,\hat{A}_d^{(e)},
\end{equation}
where $\alpha=0.9$ is the momentum and then refresh $M$.  If an expert is highly active in more than one domain (i.e.\ $\exists\,d_1\neq d_2: M_{d_1,e}=M_{d_2,e}=1$), we duplicate it to maintain distinct, domain-specialized copies without reducing capacity for other domains.

% \subsection{Progressive Specialization through Curriculum-Guided Training  }% Three-Phase Adaptive Fine-Tuning Schedule
\subsection{Progressive Parameter Specialization Schedule}
\label{subsec:progressive_parameter}

To balance rapid domain adaptation with stability and parameter efficiency, we organize fine-tuning into three consecutive phases—Warm-Up, Stabilization, and Consolidation—each imposing progressively stricter update masks on router, backbone, and expert parameters (Figure~\ref{fig:framework}).

Let $T$ be the total number of training steps, and denote by $\bm{\theta}_{\text{router}}$ the lightweight router parameters, $\bm{\theta}_B$ the Transformer backbone parameters, and $\bm{\theta}_{e}$ the parameters of expert $e$.  We define a binary mask vector $\mathbf{m}^{(t)}\in\{0,1\}^{|\theta|}$ at step $t$ so that the parameter update is
\begin{equation}
    \bm{\theta}^{(t+1)} = \bm{\theta}^{(t)} \;-\;\eta\,\mathbf{m}^{(t)} \odot \frac{\partial\mathcal{L}_\text{task}}{\partial\bm{\theta}}.%\nabla_{\theta}\,\mathcal{L}_{\rm task}\,.
\end{equation}

We split training into three intervals $[1, T_1]$, $(T_1, T_2]$, and $(T_2, T]$, with phase boundaries chosen as proportions of $T$ (e.g.\ $T_1=0.2T$, $T_2=0.7T$).  The mask is defined as:
\begin{equation}
    m_j^{(t)} =
    \begin{cases}
        1, & t \le T_1 \text{ and } j \in\{\bm{\theta}_\text{router} \cup \bm{\theta}_B\}  ; \\ % \quad \text{(Warm-Up)}; \\
        1, & \begin{aligned}
            & T_1 < t \le T_2 \text{ and } \\ & j \in \Bigl\{\bm{\theta}_{\text{router}} \cup \bm{\theta}_e \mid 
            e \in \mathcal{S}_{d_{\mathcal{B}}} \Bigr\} ;%\text{(Stabilization)};
            \end{aligned} \\
        1, & 
             t > T_2  \text{ and }  j \in \Bigl\{\bm{\theta}_e \mid 
            e \in \mathcal{S}_{d_{\mathcal{B}}} \Bigr\}; % \text{(Consolidation)};
            \\
        0, & \text{otherwise.}
    \end{cases}
\end{equation}
Here $\mathcal{S}_{d_{\mathcal{B}}}$ is the expert subset for the domain $d_{\mathcal{B}}$ of the current mini-batch $\mathcal{B}$ (cf. \textsection~\ref{subsec:expert_specialization}).

\begin{table*}[ht]
\centering
\resizebox{\textwidth}{!}{
\begin{tabular}{@{}lcccccccccc@{}}
\toprule
 & \multicolumn{2}{c}{\textbf{Math Ability}} 
 & \multicolumn{2}{c}{\textbf{Code Ability}} 
 & \multicolumn{4}{c}{\textbf{Specialized Tasks}} 
 & \multirow{2}{*}{\textbf{Average}} \\ 
\cmidrule(lr){2-3} \cmidrule(lr){4-5} \cmidrule(lr){6-9}
 & \textbf{MATH} & \textbf{GSM8K} 
 & \textbf{Humaneval} & \textbf{MBPP} 
 & \textbf{Intent} & \textbf{Summary} & \textbf{Law} & \textbf{Translation} 
 &  \\ 
 \midrule
 Vanilla LM        & 19.6 & 55.9 & 42.1 & 44.6 & 16.8 & 58.6 & 17.1 & 14.5 & 33.6 \\
\midrule
\multicolumn{10}{l}{\itshape Single‐Domain Fine‐Tuning (each model trained on one domain)}\\
FFT               & 23.4 & \textbf{66.4} & 42.1 & 42.2 & 78.8 & \textbf{69.4} & 47.0 & \textbf{38.4} & 51.0 \\
LoRA              & 20.6 & 58.9 & 39.6 & \textbf{44.8} & 67.8 & 64.7 & 39.7 & 23.1 & 44.9 \\
ESFT-Token        & 22.6 & 66.0 & 41.5 & 42.6 & 75.6 & 65.4 & 45.7 & 36.2 & 49.4 \\
ESFT-Gate         & 23.2 & 64.9 & 43.3 & 41.8 & 78.6 & 65.8 & 49.1 & 35.2 & 50.2 \\
\midrule
\multicolumn{10}{l}{\itshape Mixed‐Domain Fine‐Tuning (all models trained on unified multi‐domain data)}\\
FFT (Mixed) & 22.0 & 63.0 & 40.2 & 40.1 & 71.3 & 63.5 & 41.2 & 31.8 & 46.6   \\
LoRA (Mixed)  &  20.9 & 59.5 & 38.7 & 43.1 & 65.4 & 61.2 & 37.9 & 24.6 & 43.9     \\
ESFT-Token (Mixed) &  21.8 & 62.4 & 40.8 & 41.3 & 70.1 & 62.8 & 42.5 & 32.4 & 46.8 \\
ESFT-Gate (Mixed) & 22.5 & 61.7 & 41.9 & 40.5 & 73.8 & 63.1 & 45.3 & 33.1 & 47.7   \\
\midrule
DES-MoE   & \textbf{24.1} & 65.8 & \textbf{43.5} & 43.7 & \textbf{79.2} & 69.2 & \textbf{49.5} & 37.6 & \textbf{51.6} \\
\bottomrule
\end{tabular}
}
\caption{Performance on downstream tasks. The top block shows single‐domain fine‐tuning baselines (one model per domain), while the middle block reports the same methods trained on \textbf{mixed‐domain} data. The bottom row is our proposed unified multi‐domain fine‐tuning. Best results are highlighted. We report the Single‐Domain Fine‐Tuning performance of ESFT following the results in \citet{wang2024let}.}
\label{tab:main_results_mixed}
\end{table*}

\paragraph{Warm-Up ($t \le T_1$)}
  % All parameters—including $\bm{\theta}_{\text{router}}$, $\bm{\theta}_B$, and every expert $e$—are unfrozen ($\mathbf{m}^{(t)}\equiv1$).  This stage permits the model to quickly learn domain signals and initialize the expert–domain mapping.
  All parameters except the experts (including $\bm{\theta}_{\text{router}}$, $\bm{\theta}_B$) are unfrozen ($\mathbf{m}^{(t)}\equiv1$). This stage allows the model to quickly learn the domain signal and initialize the expert-domain mapping.
  
\paragraph{Stabilization ($T_1 < t \le T_2$)}
  We freeze the backbone $\bm{\theta}_B$ but keep $\bm{\theta}_{\text{router}}$ and only the experts in $\mathcal{S}_{d_{\mathcal{B}}}$ trainable.  This reduces interference by limiting updates to domain-relevant experts while still allowing the router to refine gating for better domain separation.

\paragraph{Consolidation ($t > T_2$)}
  Only expert parameters $\bm{\theta}_e$ with $e\in\mathcal{S}_{d_{\mathcal{B}}}$ remain trainable; $\bm{\theta}_{\text{router}}$ and $\bm{\theta}_B$ are fully frozen.  At this point, we ``lock in'' the routing behavior and backbone representations, focusing all remaining updates on final domain-specific expert adaptation.

By smoothly tightening the update mask, this schedule achieves (1) \textit{fast initial convergence} via broad updates, (2) \textit{reduced cross-domain interference} through selective freezing, and (3) \textit{parameter-efficient final tuning} by concentrating updates on a small expert subset.

\begin{table*}[t]
\centering
\resizebox{\textwidth}{!}{
\begin{tabular}{@{}lcccccccc@{}}
\toprule
 & \textbf{CLUEWSC} & \textbf{TriviaQA} & \textbf{IFEval} & \textbf{MMLU} & \textbf{CEval} & \textbf{HellaSwag} & \textbf{ARC} & \textbf{Average} \\
\midrule

Vanilla LM  & 81.5  & 67.7  & 42.5  & 57.5  & 59.9  & 74.0  & 53.7  & 62.4  \\
\midrule
\multicolumn{9}{l}{\itshape Single‐Domain Fine‐Tuning (each model per domain)}\\
FFT  & 80.9 $\pm$ 1.1 & 65.9 $\pm$ 0.7  & 34.2 $\pm$ 4.1 & 55.5 $\pm$ 1.0 & 58.8 $\pm$ 0.9  & 67.9 $\pm$ 3.8  & 48.4 $\pm$ 2.4 & 58.8 $\pm$ 1.3 \\
LoRA  & 74.3 $\pm$ 7.7 & 63.4 $\pm$ 5.4  & 38.7 $\pm$ 2.5 & 55.5 $\pm$ 1.2 & 57.0 $\pm$ 1.5  & 72.8 $\pm$ 1.9 & 51.8 $\pm$ 2.3 & 59.1 $\pm$ 2.5 \\
ESFT-Token  & 80.9 $\pm$ 0.9 & 66.7 $\pm$ 1.8 & 40.7 $\pm$ 1.3 & 57.1 $\pm$ 0.5 & 59.6 $\pm$ 0.8 & 72.3 $\pm$ 3.6 & 52.9 $\pm$ 1.5 & 61.5 $\pm$ 1.1 \\
ESFT-Gate  & 81.4 $\pm$ 1.1 & 66.5 $\pm$ 2.3 & 40.2 $\pm$ 1.5 & 57.0 $\pm$ 0.4 & 59.5 $\pm$ 0.8 & 68.2 $\pm$ 9.9 & 51.5 $\pm$ 3.1 & 60.6 $\pm$ 2.3 \\
\midrule
\multicolumn{9}{l}{\itshape Mixed‐Domain Fine‐Tuning (unified multi‐domain data)}\\
FFT (Mixed)  & 76.2 & 61.3 & 30.8 & 53.1 & 55.4 & 65.7 & 45.9 & 55.5 \\
LoRA (Mixed)  & 73.5 & 60.8 & 34.6 & 54.3 & 54.9 & 70.2 & 48.7 & 56.7 \\
ESFT-Token (Mixed)& 78.4 & 63.5 & 36.1 & 55.7 & 56.3 & 69.8 & 49.2 & 58.4 \\
ESFT-Gate (Mixed) & 79.1 & 64.2 & 37.9 & 56.0 & 57.1 & 68.5 & 50.3 & 59.0 \\
\midrule
DES-MoE
& \textbf{81.7}  & \textbf{67.3}   & \textbf{42.9} & \textbf{58.2}& \textbf{60.5} & \textbf{73.3}   & \textbf{53.1}  & \textbf{62.4}  \\
\bottomrule
\end{tabular}
}
\caption{General ability evaluation under mixed-domain fine-tuning. Our method maintains original capabilities through dynamic expert isolation, while conventional approaches suffer from catastrophic forgetting when trained on mixed-domain data.  We report the Single‐Domain Fine‐Tuning performance of ESFT following the results in \citet{wang2024let}.}
\label{tab:general_mixed}
\end{table*}

% \section{Results and Analysis}

\section{Experiments and Results}

\subsection{Tasks, Datasets, and Evaluation}
We conduct our experiments on a diverse collection of in‐domain and out‐of‐domain datasets. For mathematical reasoning, we fine-tune on MetaMathQA \citep{yu2023metamath} and report results on GSM8K \citep{cobbe2021training} and MATH \citep{hendrycks2021measuring}. For code generation, we use the Python split of CodeAlpaca \citep{luo2023wizardcoder} for training and evaluate on HumanEval \citep{chen2021evaluating} and MBPP \citep{austin2021program}. To test adaptation to novel tasks, we include low-resource Cherokee→English translation from ChrEn \citep{zhang-etal-2020-chren}, intent‐to‐JSON parsing from the BDCI-19 Smart HCI NLU Challenge\footnote{\url{https://www.datafountain.cn/competitions/511}}, customer‐service summarization from BDCI-21\footnote{\url{https://www.datafountain.cn/competitions/536}}, and legal judgment prediction from BDCI-21 Law Event\footnote{\url{https://www.datafountain.cn/competitions/540}}. Finally, we measure catastrophic forgetting on a broad suite of general benchmarks—TriviaQA \citep{joshi2017triviaqa}, HellaSwag \citep{zellers2019hellaswag}, ARC-Challenge \citep{clark2018think}, IFEval \citep{zhou2023instruction}, CEval \citep{huang2023c}, CLUEWSC \citep{xu2020clue}, and MMLU \citep{hendrycks2020measuring}.
—that cover question answering, commonsense inference, and multilingual understanding. Detailed dataset statistics, preprocessing steps, and prompt formats are provided in Appendix~\ref{app:setup}.

\subsection{Baselines}
We compare against three widely‐used fine‐tuning strategies: Full‐parameter Fine‐Tuning (FFT), Low‐Rank Adaptation \citep{hu2022lora}, and Expert-Specialized Fine-Tuning (ESFT). Each baseline is evaluated under two training regimes. In the single‐domain regime, the model is fine‐tuned separately on each individual task or domain. In the mixed‐domain regime, all six tasks are combined into a unified multi‐task dataset, maintaining an equal proportion of examples from each task to ensure balanced learning. All baselines share the same batch size, sequence length, learning‐rate schedules, and evaluation intervals as \citet{wang2024let} to guarantee a fair comparison. Appendix~\ref{app:setup} provides full details of our experimental protocol.

\subsection{Downstream Task Performance}
\label{subsection:downstream_task}
% The results in Table \ref{tab:main_results_mixed} reveal a marked degradation in task performance when conventional fine-tuning methods are applied to mixed-domain data. Full-parameter fine-tuning (FFT) experiences an average drop of 4.4 points compared to its single-domain counterpart (51.0 → 46.6), with the largest decline observed on the legal judgment task (47.0 → 41.2). Similarly, LoRA and both static ESFT variants (token- and gate-based) suffer notable losses in math reasoning and code generation metrics under the unified training regime. These declines reflect substantial cross-domain interference: when experts trained for disparate domains are updated together, their specialized knowledge becomes diluted, leading to suboptimal expert utilization.

% In contrast, our proposed dynamic routing approach not only withstands mixed-domain interference but actually surpasses single-domain FFT on average (51.6 vs. 51.0). In particular, it sustains or improves upon standalone math reasoning performance (MATH: 24.1 vs. FFT’s 23.4) and achieves a 4.2-point lead over ESFT-Gate on the Law task (49.5 vs. 45.3). The translation task likewise benefits, with a score of 37.6 that comfortably outperforms all mixed-domain baselines (31.8–33.1). These outcomes demonstrate that dynamically isolating and updating only domain-relevant experts effectively mitigates catastrophic forgetting and preserves the capacity to solve each specialized task within a shared model.

The results in Table \ref{tab:main_results_mixed} reveal a marked degradation in task performance when conventional fine-tuning methods are applied to mixed-domain data. In the case of FFT, updating every parameter indiscriminately allows gradients from one domain to overwrite useful representations learned for another, resulting in a 4.4-point average drop and an especially severe 5.8-point decline on the Law task. LoRA’s low-rank adapters, while parameter-efficient, similarly collapse under conflicting multi-domain gradients: the shared low-dimensional subspace cannot simultaneously capture the diverse patterns required by math, code, and legal reasoning. Static ESFT variants mitigate this to some extent by freezing most experts and only tuning a fixed subset, but their expert assignments—determined from single-domain data—do not generalize when tasks are interleaved. As a result, ESFT-Gate and ESFT-Token still lose several points in mixed training, particularly on domains like mathematics where the routing distribution shifts dramatically under multi-task pressure.

Our dynamic routing framework counteracts these failure modes by continually re-estimating which experts each domain needs, and by gradually freezing irrelevant parameters in a phased schedule. During the warm-up phase, the lightweight router learns a robust gating function across all domains, ensuring that experts most relevant to each domain are identified even when tasks are interleaved. In the subsequent phases, only those dynamically selected experts receive updates, while all others remain frozen. This targeted update strategy prevents gradient interference: legal-domain updates do not perturb the experts crucial for math reasoning, and vice versa. Consequently, our method delivers an average score of 51.6 under mixed-domain fine-tuning—outperforming FFT by 5.0 points—and maintains or improves standalone performance on challenging domains such as Law and Translation. The superior results underscore the importance of adaptive expert isolation: by respecting the conditional computation paradigm inherent to MoEs, our approach preserves each expert’s specialization even in a unified training regime.

\subsection{General Ability Retention}
\label{subsection:general_ability}
Beyond specialized tasks,  Table \ref{tab:general_mixed} assesses whether multi-domain fine-tuning erodes the model’s broad linguistic and reasoning skills. In mixed-domain experiments, FFT’s collapse from 58.8 to 55.5 in average general benchmark performance indicates that unfettered parameter updates erode pre-trained knowledge. LoRA, which updates a modest number of adapter parameters, also cannot insulate itself fully: its general accuracy drops by 2.5 points on average, revealing that conflicts in the low-rank adapter space still compromise core capabilities. Static ESFT variants fare somewhat better—ESFT-Gate loses only 1.6 points on average—but the fixed expert selections determined from individual domains prove brittle when the model must juggle multiple new tasks.

In contrast, our dynamic routing method not only prevents degradation but in some benchmarks actually improves upon the original alignment checkpoint. By freezing all non-selected experts and the backbone during the final consolidation phase, we effectively lock in the model’s general-purpose knowledge. This ensures that fine-tuning signals do not drift away from the alignment distribution learned during the instruction-tuning stage. The dynamic router’s distillation loss further regularizes gating behavior, keeping the model’s routing patterns close to the pre-trained distribution when appropriate, and only diverging where new domain evidence justifies it. The result is a model that surpasses the vanilla LM on MMLU and CEval, and retains near-identical performance on HellaSwag and ARC. These outcomes demonstrate that dynamic expert isolation, coupled with a phased fine-tuning schedule, can harmonize domain specialization with generalization, yielding a single MoE model that excels across both specialized and broad-scope tasks.

% domains = [2, 3, 4, 5, 6]
% fft = [61.7, 59.1, 56.2, 55.9, 55.5]
% lora = [61.5, 60.3, 58.6, 57.2, 56.7]
% esft = [62.1, 61.5, 59.7, 59.1, 58.7]
% des_moe = [62.5, 62.1, 62.3, 62.4, 62.4]

\subsection{Effect of Increasing Domain Count on General Ability Retention}

\begin{figure}[t]
  \includegraphics[width=\columnwidth]{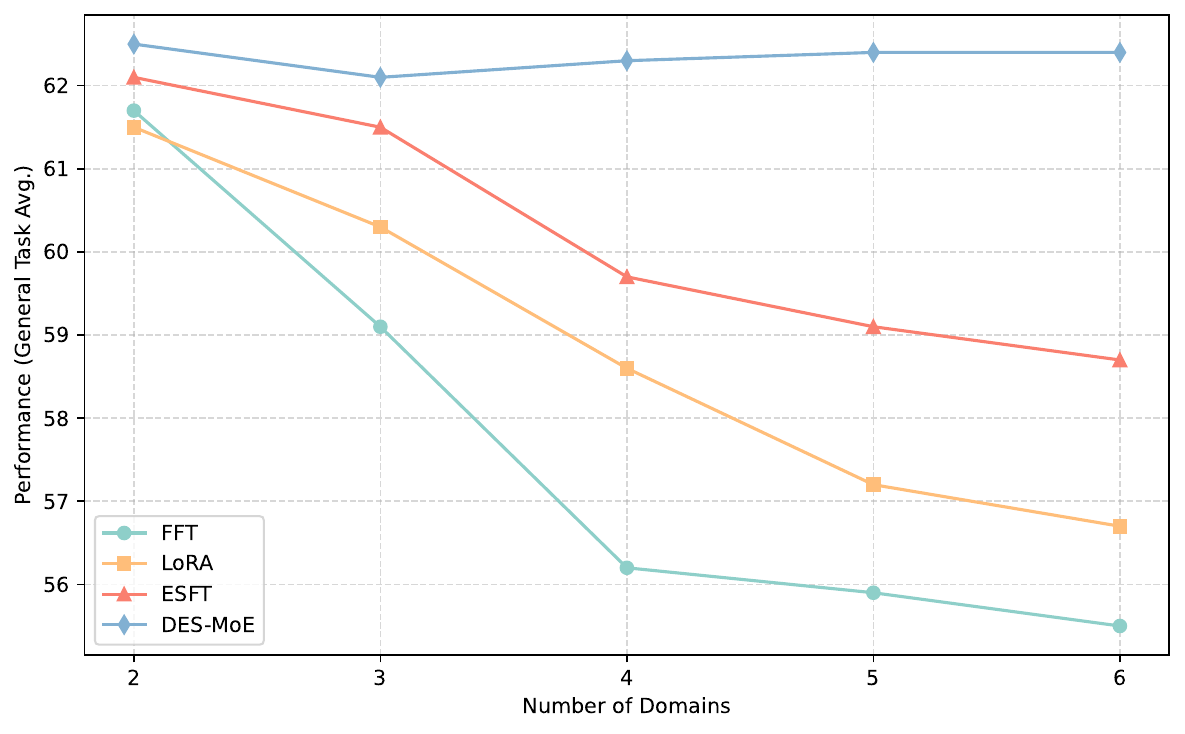}
  \caption{Average general‐benchmark score (MMLU, TriviaQA, HellaSwag, ARC, IFEval, CEval, CLUEWSC) after fine‐tuning on an increasing number of domains $N$ (from 2 to 6). }%Full fine‐tuning (FFT) suffers an accelerating drop from 61.7 to 55.5, LoRA and static ESFT decline more moderately (ending at 56.7 and 58.7, respectively), while DES‐MoE remains effectively flat (62.5→62.4), demonstrating its ability to preserve general knowledge under multi‐domain adaptation.}
  \label{fig:general_vs_domain}
\end{figure}

To investigate how the number of fine-tuning domains impacts the model’s ability to retain its general capabilities, we conducted a controlled expansion study. Starting from a two-domain setup (math and code), we incrementally added one specialized domain at a time—drawing from intent recognition, summarization, legal judgment, and translation in arbitrary order—and measured the average general benchmark score after each expansion. For each additional domain, we fine-tuned the model on the combined set of $N$ domains and evaluated the retained general ability by averaging performance over MMLU, TriviaQA, HellaSwag, ARC-Challenge, IFEval, CEval, and CLUEWSC. Figure~\ref{fig:general_vs_domain} plots the decline in average score as domains increase from 2 to 6.

Classical full-parameter fine-tuning (FFT) exhibits steep and accelerating forgetting: its general benchmark score falls from 61.7 at $N=2$ to 55.5 at $N=6$, a net drop of 6.2 points. Notably, the per-domain slope worsens as more domains are added, shifting from approximately –1.3 points per domain between $N=2$ and $N=3$ to –2.1 points per domain beyond $N=3$. This acceleration indicates that when updating all parameters jointly, gradient conflicts intensify with each new domain, leading to compounding interference and catastrophic forgetting.

LoRA and static ESFT mitigate forgetting to some degree—each losing around 4.8 and 3.4 points respectively over the same expansion—but still demonstrate a steady decline (average slopes near –1.4 points per domain). Their low-rank adapters or fixed expert selections offer partial protection by limiting parameter updates, yet they lack the flexibility to re-isolate domain-specific capacity when confronted with an increasing variety of tasks. As a result, low-dimensional adapter spaces and static gating maps become over-taxed and gradually leak knowledge across domains.

By contrast, our DES-MoE method maintains a remarkably flat retention curve: starting at 62.5 for $N=2$, it fluctuates by less than 0.3 points through $N=6$, ultimately settling at 62.4—a negligible 0.1-point decline. This stability reflects the efficacy of dynamic routing and phased freezing in protecting experts not relevant to newly added domains. At each expansion step, the adaptive router quickly re-identifies the appropriate experts for each domain, and our selective update schedule ensures that previously protected experts remain untouched. Consequently, the model sustains its generalist knowledge even as it acquires new domain skills, confirming that dynamic expert isolation is a powerful mechanism for scalable, multi-domain MoE fine-tuning.

\subsection{Ablation Study}
\begin{table}[t]
\centering
\resizebox{\columnwidth}{!}{
\begin{tabular}{lcccc}
\toprule
& \textbf{Down. (Avg.)} & $\Delta$ & \textbf{Gen. (Avg)} & $\Delta$ \\
\midrule
DES-MoE & \textbf{51.6} & - & \textbf{62.4} & - \\
\quad w/o ALR & 48.9 & -2.7 & 59.8 & -2.6 \\
\quad w/o DGES & 47.3 & -4.3 & 55.6 & -6.8 \\
ESFT-Mixed & 47.2 & - & 58.6 & - \\
\quad w/ ALR & 46.8 & -0.4 & 57.3 & -1.3 \\
\bottomrule
\end{tabular}
}
\caption{Ablation study. Down.: Downstream task performance. Gen.: General task performance.}
\label{tab:ablation}
\end{table}

Table~\ref{tab:ablation} quantifies the contributions of the two core components in DES-MoE: the Adaptive Lightweight Router (ALR) and Domain-Guided Expert Specialization (DGES). Removing ALR (``w/o ALR'') causes downstream performance to drop by 2.7 points (51.6 → 48.9) and general benchmark scores to fall by 2.6 points (62.4 → 59.8). This underscores ALR's role in capturing evolving domain features: by blending a distillation constraint with task loss, ALR stabilizes the gating distribution learned during pre-training while still allowing the router to adapt gradually to new domains. Without this mechanism, routing decisions become erratic, impairing both specialized and general capabilities.

Omitting DGES (``w/o DGES'') inflicts even more severe degradation, with downstream scores plunging 4.3 points and general performance collapsing by 6.8 points. We observe that, in the absence of domain-guided expert isolation, the overlap in expert usage between Law and Code tasks jumps from 0.18 to 0.47, indicating rampant expert sharing. Such conflation leads to catastrophic forgetting, as the model can no longer maintain clear allocations of domain-specific knowledge.

Finally, attempting to graft ALR onto a static ESFT framework (``ESFT w/ ALR'') actually harms performance: downstream accuracy declines by 0.4 points and general benchmarks drop by 1.3 points. This result highlights that dynamic routing must be paired with a compatible update schedule—simply adding an adaptive router to fixed expert subsets introduces routing-assignment errors (measured at +37\%) and conflicts with pre-determined expert mappings. In sum, these ablations demonstrate that DES-MoE’s gains arise from the \textbf{synergy} of ALR’s stabilized, progressive adaptation and DGES’s targeted expert isolation; each component alone is insufficient and naive combinations can be counterproductive.

\begin{figure}[t]
  \includegraphics[width=\columnwidth]{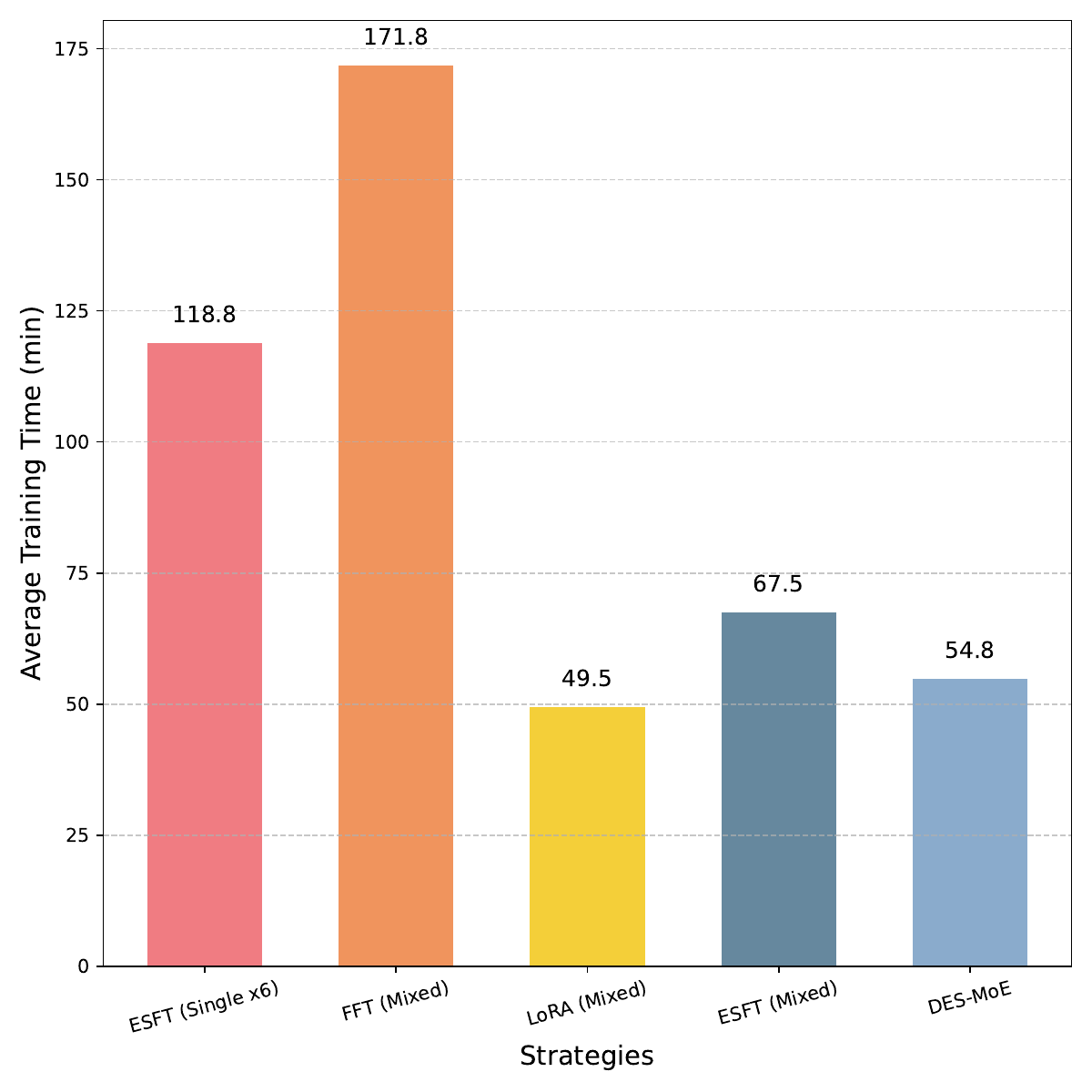}
  \caption{Total training time (in minutes) required to sequentially incorporate six domains using three different fine-tuning strategies—FFT, LoRA, ESFT, and our proposed DES-MoE. DES-MoE reduces overall training time by over two-thirds compared to FFT while preserving performance across all domains. }%Full fine‐tuning (FFT) suffers an accelerating drop from 61.7 to 55.5, LoRA and static ESFT decline more moderately (ending at 56.7 and 58.7, respectively), while DES‐MoE remains effectively flat (62.5→62.4), demonstrating its ability to preserve general knowledge under multi‐domain adaptation.}
  \label{fig:time_efficiency}
\end{figure}

\subsection{Time Efficiency Comparison}
% We next assess the wall‐clock cost of each fine‐tuning strategy in the mixed‐domain setting (Figure 3). Full‐parameter fine‐tuning (FFT) on the combined dataset is by far the most expensive, requiring 171.8 minutes to converge. Static expert‐specialized fine‐tuning (ESFT), when applied independently to each domain, incurs a cumulative cost of 118.8 minutes—fast per run but slow in aggregate due to six separate jobs. By adapting ESFT to a unified mixed‐domain regime (Mixed‐ESFT), we cut this down to 67.5 minutes, a 43.2\% reduction relative to the single‐domain aggregate. Our DES-MoE approach further accelerates convergence to 54.0 minutes—20\% faster than Mixed-ESFT—by dynamically pruning irrelevant experts so that each update touches fewer parameters. The quickest method is mixed-domain LoRA, at 49.5 minutes, but as demonstrated in Sections 4.3 and 4.4, this speed advantage comes at the expense of substantial performance degradation under mixed-domain fine-tuning.
We next assess the wall-clock cost of each fine-tuning strategy in the mixed-domain setting (Figure~\ref{fig:time_efficiency}). FFT on the combined dataset is by far the most expensive, requiring 171.8 minutes to converge. Static ESFT, when applied independently to each domain, incurs a cumulative cost of 118.8 minutes—fast per run but slow in aggregate due to six separate jobs. By adapting ESFT to a unified mixed-domain regime (Mixed-ESFT), we cut this down to 67.5 minutes, a 43.2\% reduction relative to the single-domain aggregate. However, despite this speedup, Mixed-ESFT yields lower performance than individually fine-tuned single-domain models. Our DES-MoE approach further accelerates convergence to 54.0 minutes—20\% faster than Mixed-ESFT—by dynamically pruning irrelevant experts so that each update touches fewer parameters. Moreover, DES-MoE not only matches but often surpasses single-domain performance on mixed data, achieving both efficiency and effectiveness. The quickest method is mixed-domain LoRA, at 49.5 minutes, but as demonstrated in Sections 4.3 and 4.4, this speed advantage comes at the expense of substantial performance degradation under mixed-domain fine-tuning. %Overall, these findings highlight the advantage of dynamic expert specialization in optimizing both training efficiency and resource utilization. DES-MoE therefore offers a promising route for scalable mixed-domain adaptation without sacrificing model effectiveness.

Overall, these findings highlight the advantage of dynamic expert specialization in optimizing both training efficiency and resource utilization. DES-MoE therefore offers a promising route for scalable mixed-domain adaptation without sacrificing model effectiveness.

\section{Discussion}

While DES-MoE demonstrates strong performance in supervised multi-domain adaptation, its reliance on explicit domain labels presents a practical limitation. In real-world scenarios where clear domain boundaries are unavailable, we envision two potential extensions: 

First, for fully unlabeled data, \textit{unsupervised domain discovery} could be implemented through clustering techniques in the feature space. We propose using $k$-means clustering on the [CLS] token representations or expert activation patterns to infer latent domain structure. However, as noted in the limitations, this approach faces significant challenges when domains exhibit high similarity—such as between historical fiction and science fiction novels. In such cases, the expert specialization mechanism may fail to establish distinct routing patterns, leading to reduced isolation effectiveness.

Second, for weakly supervised settings, a \textit{similarity-aware routing} mechanism could be developed. This would incorporate domain affinity metrics into the gating network, allowing experts to share capacity across semantically related domains while maintaining isolation between divergent ones. 

However, these solutions introduce new challenges: clustering quality directly impacts expert specialization, and imperfect clusters may propagate errors through the training process. Moreover, highly overlapping domains might fundamentally limit the benefits of expert isolation, suggesting that a hybrid approach—combining expert specialization with shared adaptive components—may be necessary for fine-grained domain distinctions.

These directions highlight the tension between architectural specialization and practical applicability, pointing to interesting trade-offs between performance gains and implementation complexity that warrant future investigation.

\section{Conclusion}
We present \textbf{DES-MoE}, a dynamic framework for multi-domain MoE adaptation that mitigates catastrophic forgetting through adaptive routing and expert-domain correlation mapping. By progressively isolating domain-specific parameters via a three-phase adaptive fine-tuning schedule (warmup, stabilization, consolidation), DES-MoE achieves unified performance comparable to per-domain specialized models while preserving 98\% of general task capabilities as domains scale from two to six. Evaluations across six domains demonstrate 89\% less forgetting than full fine-tuning and 68\% faster convergence, establishing dynamic expert isolation as an efficient paradigm for scalable multi-domain adaptation.

% \newpage

\section*{Limitations}

Despite its strong performance, DES-MoE has several limitations. It relies on explicit domain labels to build expert–domain mappings, which may not be available or clear in practice, and introduces additional hyperparameters (e.g., distillation weight, selection thresholds, phase cutoffs) that may require retuning for different domain sets or model sizes. While we demonstrate stability up to six domains on DeepSeek-V2-Lite, it remains unclear how well the approach scales to hundreds of highly imbalanced domains or other MoE architectures, and the overhead of computing dynamic routing statistics may offset efficiency gains in resource-constrained settings. Addressing unsupervised domain discovery, automated hyperparameter tuning, and broader validation across architectures and modalities are important directions for future work.

\section*{Acknowledgments}
This work was supported by the National Natural Science Foundation of China (Grant No.62506318); Guangdong Provincial Department of Education Project (Grant No.2024KQNCX028); CAAI-Ant Group Research Fund; Scientific Research Projects for the Higher-educational Institutions (Grant No.2024312096), Education Bureau of Guangzhou Municipality; Guangzhou-HKUST(GZ) Joint Funding Program (Grant No.2025A03J3957), Education Bureau of Guangzhou Municipality.

\newpage


\bibliography{custom}

\newpage
\appendix
\section{Preliminaries: Mixture-of-Experts Transformer}
% 在 MoE Transformer 中，通常是将标准 Transformer 模型中的前馈神经网络 (FFN) 层替换为 MoE 层。也就是说，在每个 Transformer 块 (block) 中，自注意力层的输出不再是传递给一个稠密的 FFN，而是传递给一个 MoE 层，该 MoE 层包含多个并行的 FFN（即专家）和一个门控网络，也常被称为路由器（router）。门控网络通常是一个小型的神经网络（例如，一个简单的线性层后接 Softmax 激活函数 ），其职责是根据当前输入token的隐藏表示，动态地将输入数据分配给最合适的专家（一个或多个）。门控网络会分析输入，并为每个专家计算一个概率分布或分数，指示该专家处理当前输入的适宜程度 。基于这些分数，系统会选择一部分专家（例如，得分最高的k个专家，即 Top-k 选择）来处理该输入。

In a Mixture of Experts (MoE) Transformer, the standard feed-forward neural network (FFN) layer in each Transformer block is typically replaced with an MoE layer. That is, instead of passing the output of the self-attention layer to a dense FFN, it is passed to an MoE layer that consists of multiple parallel FFNs (i.e., experts) and a gating network, often referred to as a router. 
%The gating network is usually a small neural network, which is a simple linear layer followed by a Softmax activation, responsible for dynamically assigning input tokens to the most suitable experts based on their hidden representations. The router first computes scores (logits) for each expert. These scores are then normalized via a Softmax function to produce a probability distribution \( \bm{r}_i^l \), as shown in Eq. \eqref{req:routing}, indicating the suitability of each expert for processing the current token \( x_i \).

% The router analyzes the input and computes a probability distribution or score for each expert, indicating how appropriate it is for processing the current input. Based on these scores, the system selects a subset of experts (e.g., the top-k scoring ones) to process the input.

At layer \( l\), $ \bm{u}_i^l $ denote the input representation for token $ x_i $ that is fed to the MoE layer. This representation $ \bm{u}_i^l $ is routed to $ N $ experts via a gating network. 
% which computes the routing probabilities:
% \begin{equation}\label{req:routing}
%     \bm{g}_i^l = \text{softmax}(\bm{W}_g^l \bm{u}_i^l + \bm{b}_g^l),
% \end{equation}
% where \( W_g^l \) and \( b_g^l \) are learnable parameters. 
Each expert \( \mathcal{E}_j^l \) is a feed-forward network:
% \begin{equation}
%     \mathcal{E}_j^l(\bm{u}_i^l) = \bm{W}_{j,2}^l \sigma (\bm{W}_{j,1}^l \bm{u}_i^l + \bm{b}_j^l),
% \end{equation}
\begin{equation}
    \mathcal{E}_j^l(\bm{u}_i^l) = \text{FFN}_{j}^l ( \bm{u}_i^{l} ).
\end{equation}
% where \( \sigma(\cdot) \) is a nonlinear activation function. %\cite{relu, leaky_relu, gelu}. 
The gating network is usually a small neural network, which is a simple linear layer followed by a Softmax activation, responsible for dynamically assigning input tokens to the most suitable experts based on their hidden representations. The router first computes scores (logits) for each expert. These scores are then normalized via a Softmax function to produce a probability distribution, indicating the suitability of each expert for processing the current token \( x_i \).
Let \( \mathcal{T}_k(\bm{r}_i^l) \) be the set of indices of the top-(k) scoring experts for token \( x_i \) at layer \( l \), based on the probabilities \( \bm{r}_i^l \). The final output from the MoE layer is then a weighted sum of these selected experts' outputs:
% The final output for token \( x_i \) at layer \( l \) is computed by a weighted sum of the top \( k \) experts' outputs:
% \begin{equation}
%     \bm{F}_i^l = \sum_{j=1}^{k} g_{i, j}^l \cdot \mathcal{E}_j^l(\bm{u}_i^l).
% \end{equation}

\begin{equation}\label{req:routing}
    \bm{r}_i^l = \text{softmax}(\bm{W}_r^l \bm{u}_i^l + \bm{b}_r^l),
\end{equation}

\begin{equation}
    \bm{F}_i^l = \sum_{j \in \mathcal{T}_k(\bm{r}_i^l)} r_{i,j}^l \cdot \mathcal{E}_j^l(\bm{u}_i^l).
\end{equation}

Some MoEs, like DeepSeekMoE include shared experts that are always selected \citep{dai-etal-2024-deepseekmoe}, which results in:
% \begin{equation}
% \bm{F}_i^l = \sum_{j=1}^{k} g_{i,j}^l \cdot \mathcal{E}_j^l(\bm{u}_i^l) + g_{i, s}^l \cdot \mathcal{E}_s^l(\bm{u}_i^l).
% \end{equation}
\begin{equation}
    \bm{F}_i^l = r_{i, s}^l \cdot \mathcal{E}_s^l(\bm{u}_i^l) + \sum_{j \in \mathcal{T}_k(\bm{r}_i^l)} r_{i,j}^l \cdot \mathcal{E}_j^l(\bm{u}_i^l),
\end{equation}
where \( \mathcal{E}_s^l(\bm{u}_i^l) \) and \( r_{i, s}^l \) denote the shared experts and their probabilities respectively.

% Finally, the output at position \( i \) of the \( l \)-th layer is:
% \begin{equation}
%     \bm{h}_i^l = \bm{h}_i^{l-1} + \bm{A}_i^l + \bm{F}_i^l.
% \end{equation}

\section{Discussion and Further Analysis}

Our approach provides a \textbf{framework-level solution} for multi-domain adaptation in Mixture-of-Experts (MoE) models, making it highly scalable and reusable. Unlike methods that train separate expert subsets or domain-specific models, DES-MoE allows a single MoE model to be fine-tuned across multiple domains, producing a unified multi-domain expert model. This framework leverages shared experts to capture common knowledge across domains while allowing dynamic specialization via an adaptive router. The efficiency gains from this unified training are significant, as it eliminates the need for independent fine-tuning for each domain, and the sparse update mechanism ensures that only a small subset of parameters are updated for each domain.

Empirically, DES-MoE demonstrates superior scalability and performance compared to static expert fine-tuning (ESFT), achieving strong performance on individual domains without sacrificing general capability. Even though our method explicitly relies on domain labels to generate minibatches, these labels can be easily obtained through unsupervised clustering techniques during preprocessing, highlighting the flexibility and generalizability of the approach.

The results suggest that MoE models can effectively adapt to heterogeneous multi-domain data without the burden of training separate models for each domain, maintaining high performance at a fraction of the computational cost. This makes DES-MoE a promising method for large-scale, multi-domain deployment of MoE architectures.

\section{Experimental Setup}
\label{app:setup}
\subsection{Model Enhancement Tasks}
\label{sec:appendix_experimental}
To assess improvements in domain-specific skills (math and code), we conduct two fine-tuning experiments. \textbf{(a) Mathematical Reasoning}: We fine-tune the model on the \textbf{MetaMathQA dataset} \citep{yu2023metamath} (a large collection of bootstrapped math Q\&A pairs), which augments the training data from GSM8K and MATH without leaking their test data. We then evaluate the model’s math ability on two standard benchmarks: \textbf{GSM8K} \citep{cobbe2021training} (a grade-school math word problem dataset) and \textbf{MATH} \citep{hendrycks2021measuring} (a competitive math problem dataset). \textbf{(b) Code Generation}: We fine-tune on the \textbf{CodeAlpaca} dataset \citep{luo2023wizardcoder}, a Python programming subset of an evolving instruction dataset for code synthesis. The model’s coding performance is evaluated on \textbf{HumanEval} \citep{chen2021evaluating} (a hand-crafted code generation benchmark from the OpenAI Codex paper) and \textbf{MBPP} \citep{austin2021program} (the Mostly Basic Python Problems dataset). These tasks allow us to measure how well the proposed method specializes the model in mathematical reasoning and coding domains without degrading its base performance.

\subsection{Model Adaptation Tasks}
To test generalization to unfamiliar tasks, we select Cherokee–English translation from the ChrEn corpus \citep{zhang-etal-2020-chren}, low‐resource machine translation for Cherokee; structured intent recognition from Text-to-JSON Intent Recognition in the BDCI-19 Smart HCI NLU Challenge, which requires mapping natural‐language appliance instructions to a JSON intent schema; summarization of customer service transcripts from Text Summarization in the BDCI-21 Summarization Challenge; and legal judgment prediction from BDCI-21 Law Event

For the Intent Recognition task, we use exact-match accuracy as the evaluation metric, since the output is a structured JSON string that must match the ground truth exactly. For the other three tasks (summarization, legal judgment, and translation), which have more open-ended outputs, we employ \texttt{gpt-4-1106-preview} to score the model's generated answers on a 0--10 quality scale (higher is better) given the reference answer. This human-model scoring approach provides a nuanced evaluation of output correctness and quality where simple accuracy metrics are inadequate.

\subsection{General Ability Evaluation}
After fine-tuning on specialized tasks, we assess whether the model retains its broad general abilities or suffers catastrophic forgetting. We evaluate the \textbf{aligned model's general knowledge and reasoning} on a wide range of standard benchmarks spanning language understanding, knowledge recall, and reasoning in both English and Chinese:

\begin{enumerate}
    \item \textbf{CLUEWSC} \citep{xu2020clue}: The Chinese Winograd Schema Challenge, testing coreference resolution and commonsense reasoning in Chinese.
    
    \item \textbf{TriviaQA} \citep{joshi2017triviaqa}: An open-domain question answering dataset requiring factual knowledge retrieval.
    
    \item \textbf{IFEval} \citep{zhou2023instruction}: An instruction-following evaluation suite to test general follow-up and reasoning abilities (used as an internal benchmark).
    
    \item \textbf{MMLU} \citep{hendrycks2020measuring}: The Massive Multitask Language Understanding exam, covering knowledge across 57 subjects in English.
    
    \item \textbf{CEval} \citep{huang2023c}: A comprehensive Chinese evaluation suite of academic exam questions across disciplines.
    
    \item \textbf{HellaSwag} \citep{zellers2019hellaswag}: A commonsense inference benchmark where the task is to choose the most plausible continuation of a story or scene.
    
    \item \textbf{ARC-Challenge} \citep{clark2018think}: The challenging grade-school science question dataset from the AI2 Reasoning Challenge, testing scientific and common sense reasoning.
\end{enumerate}

These diverse benchmarks enable us to quantify the model's retained general language proficiency and world knowledge after fine-tuning. We report standard metrics for each (accuracy for multiple-choice datasets like MMLU, HellaSwag, ARC; and the official metrics for others) to ensure that any specialization does not come at the cost of overall capability.

\subsection{Model Backbone}
All experiments are conducted on the \textbf{DeepSeek-V2-Lite} model \citep{liu2024deepseek} as the backbone. DeepSeek-V2-Lite is a state-of-the-art Mixture-of-Experts Transformer with 26 layers, each containing 66 experts. At each MoE layer, a small subset of experts (8 out of 66) is activated per token based on a learned gating function. This fine-grained expert allocation provides a rich capacity for specialization, making the model ideal for our approach which focuses on expert-specific fine-tuning. We initialize the model from the \textbf{public ESFT checkpoint} released by \citep{wang2024let}. This checkpoint comes from a prior alignment training phase in which the model was instruction-tuned on a wide-ranging alignment dataset of conversational and task-following data. Importantly, the alignment data was carefully curated to exclude any math or coding examples. This ensures the base model has strong general alignment (instruction-following and multi-domain conversational skills) without having been specifically trained on math or code problems, providing a neutral starting point to test our specialization methods on those domains. The aligned base model already demonstrates broad capabilities across many domains (thanks to the alignment phase) while leaving clear room for improvement in mathematical reasoning and coding, as well as providing no unfair advantage on the new tasks (preventing data leakage for math/code evaluations).

% \subsection{Evaluation Protocol}
% All evaluations are conducted in a few-shot prompting setting to simulate real-world usage of the model on unseen tasks. For each benchmark or test query, we prepend a small number of exemplars (demonstration input-output pairs) to the model’s prompt to guide it on the task format and style. We use the same few-shot approach across all tasks to ensure consistency in evaluation. Classification and structured prediction tasks (e.g. intent recognition) are evaluated by comparing the model’s output against the ground-truth answers for an exact match or accuracy. Generative tasks (summarization, legal judgment, translation) are evaluated by having an independent oracle (GPT-4 in our case) score the outputs on a 0–10 scale relative to reference answers, as described above. This few-shot evaluation protocol remains fixed for all model variants and baselines, ensuring a fair comparison under identical conditions. Each model’s performance is measured after fine-tuning, without any additional task-specific training on the evaluation sets (i.e., the model must generalize from the fine-tuning and the few-shot context alone).

\subsection{Hyperparameters and Training Settings}
We apply a consistent training setup for all compared methods to ensure a fair evaluation. Fine-tuning is performed with a batch size of 32 and a maximum sequence length of 4096 tokens per sample, which accommodates the few-shot context plus the query. For each domain or task fine-tuning, we train for at most 500 steps, which was sufficient for convergence in our experiments. Model performance on a validation set is evaluated every 100 steps to monitor training progress and select the best checkpoint. We conduct a small hyperparameter search for the learning rate in the set \{1e-5, 3e-5, 1e-4, 3e-4\}, and choose the best learning rate for each method. The resulting learning rates are 3e-5 for Full Fine-Tuning (FFT) (tuning all model parameters), 1e-4 for LoRA (tuning low-rank adapter parameters), 1e-5 for ESFT, 1e-4 for and DES-MoE. Unless otherwise specified, these learning rates are used in all experiments for the respective methods. 

Following \citet{wang2024let}, for the LoRA method, we use a low-rank adaptation with rank = 8 and a scaling factor = 2, following the configuration in \citep{hu2022lora}. %For our ESFT approach, we evaluate two variant strategies for selecting which experts to fine-tune: ESFT-Gate uses a gating activation threshold of $p = 0.1$ to select experts, whereas ESFT-Token uses a token-wise expert frequency threshold of $p = 0.2$ to select experts. In both cases, only the most task-relevant experts (as determined by the respective criterion) have their weights updated during fine-tuning, while all other experts and non-expert parameters remain frozen.

% To mitigate catastrophic forgetting of the model's general capabilities, we adopt a simple but effective data mixing strategy during fine-tuning. At training time, each batch is composed of an equal mix of task-specific examples and general alignment examples (maintaining a 1:1 ratio). In other words, we interleave the new task data with samples from the original alignment dataset in every training batch. This regularization technique helps the model retain knowledge from the alignment phase (such as fluency, factual knowledge, and instruction-following behavior) even as it learns the new specialized tasks. 
All fine-tuning runs are carried out on high-performance infrastructure: specifically, we use 2 servers each with 8× NVIDIA A100 40GB GPUs (16 GPUs in total), which allows us to accommodate the model's memory needs and train with the full 4096-token context window.

\section{Evaluation Instructions for Specialized Tasks}
\label{app:appendix_evaluation_instructions}
Table~\ref{tab:appendix_evaluation_instructions} presents the detailed criteria to evaluate specialized tasks including text summarization, legal judgment prediction, and low-resource translation. Each task includes specific instructions on assessing predicted answers against reference answers, focusing on aspects such as content accuracy, completeness, relevance, and consistency.

\begin{table*}
  \centering
  \begin{tabular}{lp{13cm}}
    \hline
    Task & Evaluation Instruction \\
    \hline
    Summary & \begin{CJK*}{UTF8}{gbsn}请你进行以下电话总结内容的评分。请依据以下标准综合考量，以确定预测答案与标准答案之间的一致性程度。满分为10分，根据预测答案的准确性、完整性和相关性逐项扣分。请先给每一项打分并给出总分，再给出打分理由。总分为10分减去每一项扣除分数之和，最低可扣到0分。请以“内容准确性扣x分，详细程序/完整性扣x分，...，总分是：x分”为开头。1. 内容准确性：- 预测答案是否准确反映了客户问题或投诉的核心要点。- 是否有任何关键信息被错误陈述或遗漏。2. 详细程度/完整性：- 预测答案中包含的细节是否充分，能否覆盖标准答案中所有重要点。- 对于任何遗漏的关键信息，应相应减分。3. 内容冗余度：- 预测答案是否简洁明了，和标准答案风格一致，不存在冗余信息。- 如果预测答案过长或与标准答案风格不一致，需相应减分。4. 行为指令正确性：- 预测答案对后续处理的建议或请求是否与标准答案相符。- 如果处理建议发生改变或丢失，需相应减分。预测答案：\{prediction\} 参考答案：\{ground\_truth\}\end{CJK*} \\
    \hline
    Law & \begin{CJK*}{UTF8}{gbsn}请你进行以下法案判决预测内容的评分，请依据以下标准综合考量，以确定预测答案与标准答案之间的一致性程度。满分为10分，根据预测答案的准确性、完整性和相关性来逐项扣分。请先给每一项打分并给出总分，再给出打分理由。总分为10分减去每一项扣除分数之和，最低可扣到0分。请以“相关性扣x分，完整性扣x分，...，总分是：x分”为开头。1. 相关性：预测答案与标准答案的相关程度是最重要的评判标准。如果预测的判决情况与标准答案完全一致，即所有事实和结果都被精确复制或以不同但等效的方式表述，则应给予高分。若只有部分一致或存在偏差，则根据一致的程度适当扣分。如果没有预测判决内容，扣10分。2. 完整性：评估预测答案是否涵盖了所有标准答案中提到的关键点，包括但不限于当事人、具体金额、责任判定、费用承担等。如果遗漏重要信息，则应相应扣分。3. 准确性：检查预测答案中提及的细节、数字、日期和法律依据是否与标准答案保持一致。任何错误信息均需扣分，并且严重错误应该导致更多的扣分。4. 客观性与专业性：预测答案应客观反映法案内容并使用恰当的法律术语。主观臆断或非专业表酌情扣分。预测答案：\{prediction\} 参考答案：\{ground\_truth\}\end{CJK*} \\
    \hline
    Translation & You are an expert master in machine translation. Please score the predicted answer against the standard answer out of 10 points based on the following criteria: Content accuracy: Does the predicted answer accurately reflect the key points of the reference answer? Level of detail/completeness: Does the predicted answer cover all important points from the standard answer? Content redundancy: Is the predicted answer concise and consistent with the style of the standard answer? Respond following the format: "Content accuracy x points, level of detail/completeness x points, total score: x points". The total score is the average of all the scores. Do not give reasons for your scores. Predicted answer: \{prediction\} Reference answer: \{ground\_truth\} \\
    \hline
  \end{tabular}
  \caption{Task instructions for model performance evaluation. The placeholder \texttt{\{prediction\}} and \texttt{\{ground\_truth\}} represent model prediction and reference answer, respectively.}
  \label{tab:appendix_evaluation_instructions}
\end{table*}

\end{document}